# Forecasting Model for Crude Oil Price Using Artificial Neural Networks and Commodity Futures Prices

Siddhivinayak Kulkarni
Graduate School of Information Technology and
Mathematical Sciences
University of Ballarat
Ballarat, Australia
S.Kulkarni@ballarat.edu.au

Imad Haidar
Graduate School of Information Technology and
Mathematical Sciences
University of Ballarat
Ballarat, Australia
I.Haidar@students.ballarat.edu.au

*Abstract*— **This paper presents a model based on multilayer feedforward neural network to forecast crude oil spot price direction in the short-term, up to three days ahead. A great deal of attention was paid on finding the optimal ANN model structure. In addition, several methods of data pre-processing were tested. Our approach is to create a benchmark based on lagged value of pre-processed spot price, then add pre-processed futures prices for 1, 2, 3,and four months to maturity, one by one and also altogether. The results on the benchmark suggest that a dynamic model of 13 lags is the optimal to forecast spot price direction for the short-term. Further, the forecast accuracy of the direction of the market was 78%, 66%, and 53% for one, two, and three days in future conclusively. For all the experiments, that include futures data as an input, the results show that on the short-term, futures prices do hold new information on the spot price direction. The results obtained will generate comprehensive understanding of the crude oil dynamic which help investors and individuals for risk managements.**

## I. INTRODUCTION

Crude oil is a key commodity for global economy. As a matter of fact, it is a vital component for the economic development and growth for industrialized and developing countries in a likely manner. Moreover, political events, extreme weather, speculation in financial market, amongst others are major characteristics of crude oil market which increase the level of price volatility in the oil markets. The effect of oil price fluctuation extends to reach large number of goods and services which have direct impact on the economy as well as the communities. Therefore, to reduce the negative impact of the price fluctuations, it is very important to forecast the price direction. Unfortunately, fundamental variables such as oil supply, demand inventory, GDP[1] are not available on daily frequency which adds additional difficultly to the prediction.

Although it could be argued that the era of oil is about to be over. However, some studies have indicated that the global demand will continue to rise for the long-term despite the fact that oil demands from OECD countries have decreased, However, the overall demand for oil has increased and this to a large extent due to the increasing demands of non OECD countries, especially China [1]. Furthermore, the fact that significant amount of oil come from the unstable Middle East means more price fluctuations are expected. Therefore, forecasting oil price direction is very useful for market traders and for individuals.

In this paper we present a ANN model for crude oil price prediction for the short-term. In addition we test whether crude oil future prices[2] contain newer information about spot price direction on the short-term.

This paper proceeds as follows, Section 2 represents short literature review, Section 3 presents data description, and pre-processing along with our methodology. Section 4 details the results and discussion, and finally the paper is concluded in Section 5.

## II. LITERATURE REVIEW

There has been an avalanche of studies since the oil shock of 1973-74 on forecasting crude oil price. In this section we present a very brief review of the related and recent studies.

Moshiri and Foroutan [3] examined the chaos and nonlinearity in crude oil futures prices. Performing several statistical and econometrical tests led them to conclude that futures prices time series is stochastic, and nonlinear. Moreover, the authors compared linear and nonlinear models for forecasting crude oil futures prices, namely, they compared ARMA and GARCH, to ANN, and found that ANN is superior and produces a statistically significant forecast. However, in our opinion two points can be made regarding this study. First, the authors used raw data as input to ANN. Second, they trained the network with rather old information (1983 to 2000). Later in this paper we explain the short-fall of these two points.

Wang et al [4] present a hybrid methodology to forecast crude oil monthly prices. The model consists of combination of three separate components, Web mining from which the authors extract rule based system, in addition ANN, and ARIMA models. These three components work disjointedly, and then intergraded together to get the final results. They claimed that nonlinear integration of these three models has outperformed any single one. However, there are several issues in this system. For example, the rule base system of the text mining model [3] depends on the knowledge base which developed by human experts. This process is not only controversial, but also unreliable, because experts opinions vary on the same problem. Moreover, neither the rules nor the knowledge base was made available to the public. Wen et al [6] proposed SVM model for monthly crude oil prices, the authors claim that SVM is outperforming MLP and ARIMA for out of sample. Nevertheless, both studies used monthly price which limit the data sample significantly. In addition, the sample includes rather old data from 1970.

---

[2] Future contracts are defined as: 'A firm commitment to make or accept delivery of specified quantity and quality of commodity during a specific month in the future at price agreed upon at the time the commitment is made' [2] p. 6.

The relation between futures prices and spot price has been the centre of attention for a large number of studies, and the literature is rich with several studies covering a range of aspects with respect to this relationship. Lead-lag, efficiency, prediction amongst other factors, are the most studied areas in futures-spot literature.

The idea of using commodity futures price to predict spot price is based on the assumption that the futures price reacts faster to the new information entering the market than spot price. According to Silvapulle, and Moosa [7] trading in the futures market has many advantages, such as low transaction cost, high liquidity, and low cash in up-front, among others. This makes it much more attractive for investors to react for new information than taking position in the spot market. This argument applies for most of the commodity listed in the financial markets; however, it is more relevant to the energy markets. The reason for this is, when new information related to the oil market is introduced, investors have two options, either to take a position (buy or sell) in spot or in the futures market. In most of the cases, taking a position in spot market is not the best way for reacting to the new information, because it requires high transaction costs, storage costs, and delivery costs etc. This is especially true, if investors are not interested in the commodity itself rather they are hedging for another commodity, or simply just investing in the market in hope of arbitrage opportunity i.e. speculation. In this context, futures market is much more attractive place for an investor to react to new information for the reasons discussed above [7]. However, Brooks, et al [8] examined the lead-lag relation on high frequency data 10 min, for FTSE index. The authors claim the lead-lag[4] relation could only hold for no more than half hour, and their results confirm that changes in futures prices help predicting the changes in spot price.

On the other hand, an early study by Bopp and Sitzer [9] tested whether futures prices are good predictors for cash price in the future for the heating oil market. In attempt to answer if futures prices has the capacity to improve forecasting ability of econometrical models. The results showed that only futures contract 1 and 2 months to maturity are statistically significant for cash price forecast. In other words contain new information. Similar results were obtained by Chan [10] who studied the lad-lag relationship for S&P500. He concluded that futures market is the main source of information on market wide level, while cash market is the main source of firm specific information. Moreover, Silvapulle, and Moosa [7] results on the lead-lag of the oil market suggested that the pattern of lead lags is not constant and changeable over time. In a related study by Coppola [11] on crude oil market; he found some evidence that futures contracts are able to reflect the information about the spot price future. Abosedra & Baghestani [12] tested monthly future prices for long term-forecast. The results showed that only future 1-and 12 months ahead produced significant forecast and could be useful for policy making purposes.

Although the body of oil literature is substantial, there is still a great deal of inconsistency in the findings. This is particularly the case in the relation between spot prices and futures price. While most studies agree on the importance of futures prices for financial markets, only a few studies, if any, agree on how, and why it is important. Furthermore, the vast majority of the literature is based on analytical models. A major shortfall of econometrical model is making strong assumption about the problem. This means if the assumptions are not correct; the model could generate misleading results. Furthermore, a recent survey by Labonte [13] concludes that one of the most common cavities of econometrical models is omitted variable and in some cases structural misspecification.

### III. RESEARCH METHODOLGY

By viewing the market as a model that takes historical and current information as an input then market participants react to this information based on their understanding, positions, speculations, analysis etc. the aggregation of market participants' activities will finally translate into output or closing price. In order to imitate the market, a model needs to take a subset of the information available, and try map it to the desirable target, then greater a forecast (figure 3.1), of course with a certain degree of accuracy i.e. with an error [14].

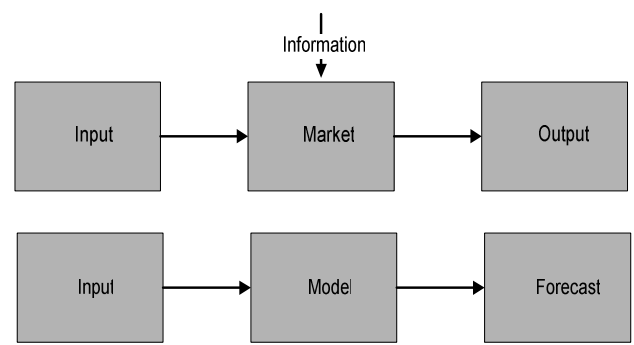

**Figure 1 model selection to mimic the market [14] p. 223**

In this context ANN is selected as a mapping model, and viewed as nonparametric, nonlinear, assumption free model [15]. This means it does not make *a priori* assumption about the problem; rather it lets the data speak for itself [14]. Furthermore, it has been proven that feedforward network with nonlinear function is able to approximate any continuous function [16], and has been around for a while now and successfully used in several studies for variety of problems including crude oil price forecast.

---

[3] For a survey of text mining for financial prediction see [5]
[4] Lead-lag refers to a type of relation between futures prices and spot price for a given asset or commodities; in which the futures prices lead spot price, while the spot price lag behind the futures.

The methodology of this study is based on three layers feedforward[5] network with backpropagation algorithm. The goal is to forecast crude oil prices for the short-term, and test whether oil futures prices contain newer information about the direction of spot price in the near futures, three days ahead. In addition, whether information in futures price integrated with spot price will lead to better forecast accuracy, the overall strategy is, creating a benchmark based on the current and past information embedded in crude oil spot price solely, using three layer feedforward ANN. Once this benchmark is created futures prices are added and performance is measured. In order to do so effectively, attention was paid to finding optimal ANN model.

*A. Design Considerations*

In the broadest sense, there are three main requirements for any successful ANN model [14]:
- Convergence, or in-sample accuracy.
- Generalization, the ability of the model to perform with new data.
- Satiability, consistency of the network output.

To insure the above points are successfully met, a large number of considerations need to be taken into account, the size and frequency of the data, network architect, the number of hidden neurons, activation function optimization methods, amongst other. Although design guide lines, and some rules of thump do exists. However, there is no evidence that any of these rules should work for a given problem. Therefore designing neural networks could be challenging task. Figure 3.2 shows the main steps to design neural network for prediction task. We chose to flow a systematic manual approach for finding the optimal network design for this problem.

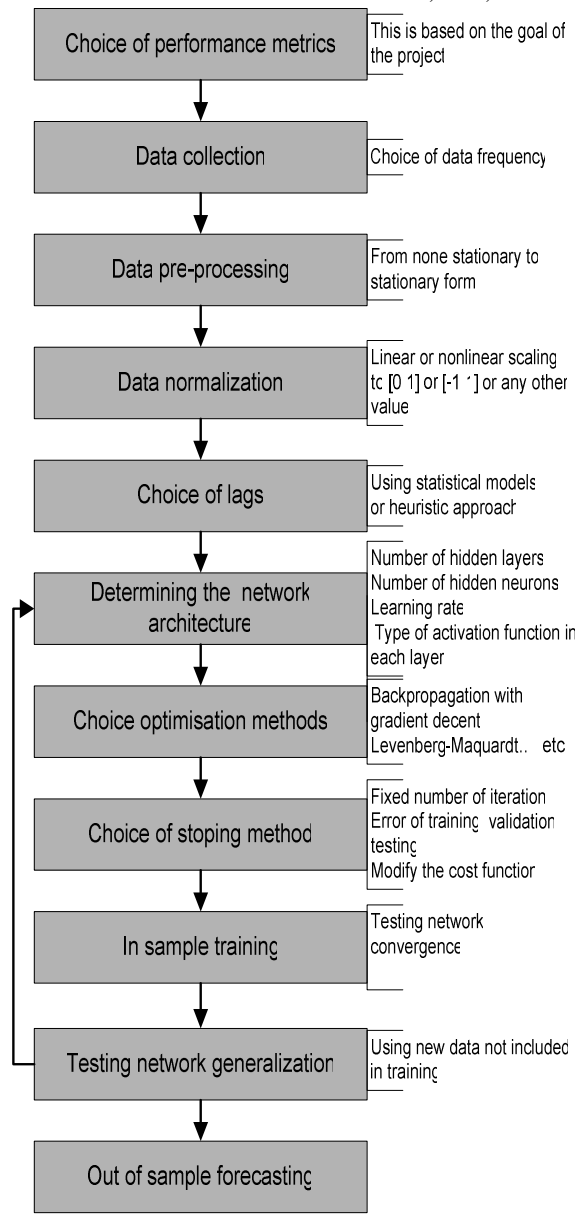

*Figure 3.2: A flow chart of the main steps for developing ANN models [17] p. 48 (modified)*

*B. Choice of Performance Metrics*

The first step in network design is to choose how to measure the performance of the system. For this study the ultimate goal is to provide a risk management tool. Therefore, our emphasis is on successfully predicting the direction of the price, rather than the magnitude, since it is very difficult, if not impossible, to correctly predict the magnitude of the price for financial data. Besides, our aim is not profitability rather risk management, hence predicting the direction is sufficient to fulfil this goal. The success ratio for direction prediction (or the hit rate) was considered [18].

---

[5] Recurrent networks were also tested, but they made no significant difference in addition to this they were computing extensive.

$$h = \frac{1}{n}\sum_{n=1}^{n} z \quad (3.1)$$

$z = 1$ if $x_{t+1}.o_{t+1} > 0$, and 0 otherwise.

where: $n$ is the sample size $x_{t+1}, o_{t+1}$ are the value of the target and the output at time $t+1$ consecutively. In addition to this the root mean squared error was also used. The RMSE is by far the most used metric for ANN performance regardless of the network goal. Furthermore, the correlation coefficient R and $R^2$ was also used; as a measure of the linear correlation between the forecasted value and the actual one [14]. Mean squared error, means absolute error, and sum squared error was also calculated. Finally, the information coefficient given by equation 3.2 was used.

$$Ic = \frac{\sqrt{\sum_{t=1}^{n}(y_t - x_t)^2}}{\sqrt{\sum_{t=1}^{n}(x_t - x_{t-1})^2}} \quad (3.2)$$

where: $y$ is the predicted value, and $x$ is the actual value. This ratio provides an indication of the prediction compared to the trivial predictor based on the random walk [14]. Where $Ic \geq 1$ indicate poor prediction, and $Ic < 1$ means the prediction is better prediction than the random walk.

*C. Data Collection*

Important point for networks design is determining the data frequency and data size. This is mainly depending on the final goal. For short-term forecast, high frequency data is preferred i.e. intraday or daily data. However, this data is not always available and in most of the case is very costly to purchase. On the other hand, weekly, and monthly data are preferred for other forecasting horizon because it is less noisy. Equally important consideration is the length of the data. Generally, when dealing with ANN the more data points, the better the network generalization. Nonetheless this is not necessarily the case when dealing with financial or economical time series. As economic conditions change over time, hence non-current (old information) could affect prediction results negatively. Because training the network with irrelevant information to the current conditions could result in a poor model generalization [18, 19].

In this study five time series are used, West Texas Intermediate (WTI) light sweet crude oil spot price and futures contracts traded at NYMEX. Futures data include four contracts 1, 2, 3, 4 months to maturity. The data frequency is daily closing price; from Sep 1996 to Aug 2007, it includes 2705 data points for each time series. All data sets retrieved from US Department of energy: Energy Information Administration web site: http://www.eia.doe.gov/ [20]. The data was divided into training and testing sets, we use 90% of the data for training and 10% for out-of-sample testing (approximately one financial year).

*D. Pre-processing and Normalization*

Pre-processing the raw data is a very sensitive issue. Although in one hand it is desirable to use raw data, since pre-processing the data could destroy the structure inbuilt only in the original time series [15, 21]. However, on the other hand in order to fit a statistically sound model, time series need to be stationary. A weakly stationary process should have a constant mean, variance, and autocovariance in its first and second momentum. Constant autocovariance means that the covariance of any sequential values is the same for stationary series [22].

For statistical point of view it is essential to have a weak stationary time when modelling for several reasons. First and foremost, the behaviour of stationary time series differs from non-stationary one. As for stationary series a shock will have less influence over time while for non-stationary the influence of shock remains for longer time steps [22]. Furthermore, using non-stationary data could lead to misleading results or so called 'spurious regressions' [14, 22]. For ANN model using non-stationary data makes it easier for the network to approximate the general characteristics of the data rather than the actual relationship [14]. Unfortunately, most of economical and financial series are non-stationary, which make transformation into stationary form essential. Several methods of transformation do exist, logarithmic difference, logarithmic return, amongst other. Neuneier & Zimmermann [23] and Gorthmann [24] suggested the use of the combination of first order relative change (equation 3.3) to represent the change in direction *momentum,* and the second order relative change *force* (equation 3.4) to represent turning point of the tie series as network input.

$$y_t = \left(\frac{x_t - x_{t-n}}{x_{t-n}}\right) \quad (3.3)$$

$$y_t = \left(\frac{x_t - 2x_{t-n} + x_{t-2n}}{x_{t-n}}\right) \quad (3.4)$$

where: $x$ is the original time series and $y$ is the transformed of $x$ and $n$ is the forecast horizon.
As the authors did not explain the pre-processing of the target we chose to test combination of these methods in addition to 3 day moving average filter on the raw data then transform it into relative change.

Another important issue is network normalization to transfer the data to fit within the limit of transfer function. A linear normalization method[6] is given by equation 3.5 to transfer the data to fit between [-1, 1] [18].

$$y_{z,t} = 2 \cdot \frac{x_{z,t} - \min(x_z)}{\max(x_z) - \min(x_z)} - 1 \quad (3.5)$$

*E. Network Architecture*

Two types of network architects were considered, a recurrent network (Elman) and multilayer feedforward network. Another important design consideration is the choice of the activation function[7] because it serves as the nonlinear part of the model. The most widely used activation functions for financial application in the hidden layer are the sigmoid functions and the hyperbolic tangent, also known as the symmetrical sigmoid function. [25]. According to McNelis 2005 [18] sigmoid functions are mostly used in financial applications as transfer functions because of its *threshold behaviour* which describes most of financial and economical series. However, the recent trend in financial application is toward the hyperbolic tangent.

The most vital consideration, however, is finding the optimal number of hidden neurons since this controls the number of free parameters in the model. This is especially important because, too many neurons result in over fitting while too few could lead to under fitting. Therefore, we used manual systematic approach; for each network 1 to 10 neurons are considered. Each network was run for minimum of three times after being re-initiated with different set of weights. Finally, other parameters such as learning rate, training time, optimization algorithm were selected base on experiments.

## IV. EXPERIMENTAL RESULTS AND ANALYSIS

*A. ANN Model*

Two networks topology was compared a fully connected feedforward networks with one, and two, hidden layers to recurrent network (Elman) with one hidden layer. In theory recurrent networks have an advantage over feedforward by modelling dynamic relationship since the output of the neuron is function to the previous input as well as to the current one, nonetheless, by adding number of lagged value of the inputs feedforward can model such dynamic relationship as well [17]. Both types has produced comparable results, however, recurrent networks took very long time to converge (around 8 hours for 100 iterations on stand alone machine). In addition to this since Elman networks keep different numbers of previous steps in the memory layer the results vary each time we re-train which violate the condition of constancy. Moreover, the recurrent networks need more hidden neurons to converge because of the memory layer. Hence, we focus our interest on modelling with feedforward network. The Choice of activation function, learning rate, was determined by experiments. The optimisation algorithms[8] also was compared and Leveberg-Maquardt was chosen as it approximates the second order, which leads to fast convergence, and higher hit rate compared to first order algorithms like gradient decent.

TABLE I. HIT RATE AND RMSE FOR IN-SAMPLE AND OUT-OF-SAMPLE

| Lag | Hit Rate (%) | | RMSE | |
|---|---|---|---|---|
| | In sample | Out of sample | In sample | Out of sample |
| 1 | 65.89 | 64.44 | 0.0313 | 0.0637 |
| 2 | 67.79 | 71.11 | 0.0292 | 0.0209 |
| 3 | 69.97 | 69.77 | 0.0276 | 0.0204 |
| 4 | 70.09 | 70.22 | 0.0273 | 0.0201 |
| 5 | 70.98 | 71.55 | 0.0266 | 0.0199 |
| 6 | 71.10 | 70.66 | 0.0264 | 0.0203 |
| 7 | 71.30 | 70.66 | 0.0260 | 0.0202 |
| 8 | 72.47 | 71.55 | 0.0256 | 0.0203 |
| 9 | 72.27 | 73.77 | 0.0254 | 0.0199 |
| 10 | 72.80 | 72.88 | 0.0248 | 0.0206 |
| 12 | 72.92 | 72.88 | 0.0244 | 0.0206 |
| 14 | 73.80 | 73.77 | 0.0233 | 0.0208 |
| 16* | 73.89 | 72 | 0.0233 | 0.0204 |
| 18* | 75.42 | 72.44 | 0.0229 | 0.0202 |
| 20* | 74.89 | 72 | 0.0242 | 0.0210 |

*= model is not stable anymore

In general, as can be seen from Table 4.3 the results were ranging around 50% which is not unusual for noisy data. For noisy data the expectation of the output to be correctly predicted is $\frac{1}{2} + \varepsilon$ of the time [14]. This is mainly for two reasons, the first is the noise in the data[9], and the seconds is the limitation of the data sample. Solving noise issue can be done through noise filter such as moving average, while solving the data limitation or lack of information is a more a difficult task, although, in general more feature need to be added to subsidise for the missing information, alternatively if there is any a priori knowledge about the function of the input/ output this can be modelled and as hint to improve the learning process [14]. Unfortunately, this is not the case of oil time sires, therefore we have to rely on filtering the noise. On the other hand, data transformation using eq. 3.4 has generated better results for in and out of sample, as eq. 3.4 contain two steps differencing. Nonetheless, combinations of these transformation methods were tested as well using one lag each. The best combination which outperformed all other options in is momentum eq. 3.3 and force eq. 3.4 as input, and force solely as target. Following this further, only this combination was tested for multiple lags (up to 12 lag form each equation). The hit rate performance was improved about 8% compared to transformation by eq 3.4 as input and target

---

[6] A nonlinear normalization methods ([18] p. 64,65) were also tested but did not improve the results.
[7] See Duch and Jankowski [26] for a survey of different activation functions.
[8] The convergence time was also tested when comparing among optimisation algorithms, as some algorithms need longer time to converge.
[9] For more information on the effect of noisy data on neural network performance, see Refenes 1995 p.60 and p. 223,224.

alone, while comparing to eq 3.3 the improvement was substantial.

Finally, the best performance which is considered as candidate for the benchmark is a combination of 7 lags of momentum and 7 lags of force (eq 3.3, 3.4) in the input, and 1 lag of force as target. Furthermore, the network architecture is three layers feedforward with 8 neurons in the hidden layer. The network was trained for 1000 iterations or until one of the stopping criteria is met. The learning rate is 0.01 and training algorithm is Levenberg-Marquardt. Table 4.5 summarizes the benchmark performance averaged over 5 trials.

TABLE II. SUMMARY OF PERFORMANCE OF CANDIDATE BENCHMARK

| Metric | Hit Rate | RMSE | MSE | MAE |
|---|---|---|---|---|
| In-Sample | 74.93 | 0.02312 | 0.00052 | 0.01724 |
| Out-of-Sample | 76 | 0.01922 | 0.00038 | 0.01502 |

In addition, 3 days moving average was applied on the raw data, then data transformed into relative change. Table 4.6 presents the results at different lags after applying the moving average filter.

TABLE III. HIT RATE AND RMSE INPUT 3 DAY FOR IN-SAMPLE AND OUT-OF-SAMPLE

| Lag | Hit Rate (%) | | RMSE | |
|---|---|---|---|---|
| | In sample | Out of sample | In sample | Out of sample |
| 1 | 72.77 | 73.75 | 0.0108 | 0.0079 |
| 2 | 72.88 | 74.01 | 0.0107 | 0.0080 |
| 3 | 73.74 | 74.54 | 0.0104 | 0.0077 |
| 4 | 75.42 | 76.37 | 0.0099 | 0.0073 |
| 5 | 76.02 | 76.90 | 0.0096 | 0.0074 |
| 6 | 76.16 | 77.16 | 0.0095 | 0.0072 |
| 7 | 77.25 | 77.11 | 0.0085 | 0.0077 |
| 8 | 78.01 | 75.59 | 0.0091 | 0.0070 |
| 9 | 78.23 | 76.77 | 0.0088 | 0.0070 |
| 10 | 77.78 | 78.08 | 0.0089 | 0.0069 |
| 11 | 78.03 | 76.37 | 0.0086 | 0.0071 |
| 12 | 77.97 | 77.95 | 0.0087 | 0.0070 |
| 13 | 79.45 | 79.79 | 0.0083 | 0.0068 |
| 14 | 79.39 | 77.42 | 0.0083 | 0.0072 |
| 15 | 79.75 | 79.11 | 0.0078 | 0.0073 |
| 16 | 79.77 | 79 | 0.0081 | 0.0071 |
| 17 | 79.45 | 78.34 | 0.0080 | 0.0069 |
| 18 | 80.40 | 78.87 | 0.0076 | 0.0133 |
| 19 | 80.95 | 77.16 | 0.0076 | 0.0075 |
| 20 | 81.38 | 77.55 | 0.0074 | 0.0074 |

Clearly, 3 days simple moving average has improved the results significantly, for in and out of sample. The results above suggest that 13 lags are optimal, it produced relatively high hit rate and the hit rate was the same for in-sample as well as for out-of-sample. In addition, the RMSE was also low. Beside, the model was stable as we tested it for 5 different trails with different sets of weights. The number of lags is extremely important as it explains the dynamics of the system. Static model which includes one lag of the dependent variable tests only the simultaneous relationship between the variable [22]. This is a very important point in financial model; as market participant either under react or over react to new news, hence after new news introduced the price might go up (down) sharply and settles down after a time (lags) therefore, including one lag is usually not enough and could generate misleading results [22].

### B. Futures Prices

In order for futures contracts to predict spot price directions, information embedded in futures data should improve the overall results. Otherwise, the assumption that futures prices contain newer information about of spot price direction cannot be accepted.

We start measuring how much information could be extracted from each futures contract about the direction of spot price next day. The moving average and relative change transformation was applied to all futures contracts and each contract was presented to the network as input while the spot price was in the same transformation.

TABLE IV. FUTURES 1 PERFORMANCE AT DIFFERENT LAGS

| Lag | Hit Rate (%) | | RMSE | |
|---|---|---|---|---|
| | In sample | Out of sample | In sample | Out of sample |
| 1 | 71.72 | 73.55 | 0.0112 | 0.0079 |
| 2 | 71.92 | 73.80 | 0.0109 | 0.0081 |
| 3 | 71.42 | 71.96 | 0.0108 | 0.0079 |
| 4 | 73.65 | 75.15 | 0.0104 | 0.0075 |
| 5 | 74.37 | 74.91 | 0.0094 | 0.0085 |
| 6 | 74 | 76.51 | 0.0102 | 0.0074 |
| 7 | 75.23 | 77.61 | 0.0099 | 0.0073 |
| 8 | 74.99 | 75.26 | 0.0090 | 0.0081 |
| 9 | 75.40 | 75.52 | 0.0096 | 0.0085 |
| 10 | 75.64 | 77.49 | 0.0095 | 0.0072 |
| 11 | 76 | 76.51 | 0.0095 | 0.0073 |
| 12 | 75.82 | 77.24 | 0.0095 | 0.0070 |
| 13 | 76.35 | 76.01 | 0.0093 | 0.0073 |
| 14 | 76.91 | 75.89 | 0.0092 | 0.0074 |
| 15 | 76.78 | 75.52 | 0.0092 | 0.0074 |
| 16 | 76.84 | 78.11 | 0.0091 | 0.0072 |
| 17 | 77.94 | 75.52 | 0.0088 | 0.0076 |
| 18 | 78.09 | 76.63 | 0.0087 | 0.0079 |
| 19 | 77.66 | 77.24 | 0.0087 | 0.0078 |
| 20 | 78.87 | 76.14 | 0.0086 | 0.0078 |

TABLE V. FUTURES 2 PERFORMANCE AT DIFFERENT LAGS

| Lag | Hit Rate (%) | | RMSE | |
|---|---|---|---|---|
| | In sample | Out of sample | In sample | Out of sample |
| 1 | 70.96 | 71.46 | 0.0115 | 0.0083 |
| 2 | 70.80 | 72.32 | 0.0112 | 0.0086 |
| 3 | 70.93 | 74.29 | 0.0111 | 0.0083 |
| 4 | 72.87 | 74.54 | 0.0107 | 0.0080 |
| 5 | 73.20 | 74.54 | 0.0106 | 0.0080 |
| 6 | 73.13 | 74.54 | 0.0105 | 0.0079 |
| 7 | 74.44 | 74.29 | 0.0103 | 0.0077 |
| 8 | 75.22 | 74.17 | 0.0100 | 0.0078 |
| 9 | 74.84 | 72.94 | 0.0101 | 0.0079 |
| 10 | 75.12 | 75.52 | 0.0098 | 0.1078 |
| 11 | 75.71 | 75.28 | 0.0097 | 0.0077 |
| 12 | 76.21 | 76.75 | 0.0097 | 0.0181 |
| 13 | 76.44 | 76.63 | 0.0094 | 0.0078 |
| 14 | 76.64 | 75.65 | 0.0096 | 0.0083 |
| 15 | 76.74 | 75.15 | 0.0094 | 0.0521 |
| 16 | 76.74 | 74.78 | 0.0091 | 0.0086 |
| 17 | 75.52 | 73.75 | 0.0109 | 0.0085 |
| 18 | 77.64 | 74.29 | 0.0091 | 0.0082 |
| 19 | 77.35 | 74.42 | 0.0089 | 0.0085 |
| 20 | 77.28 | 73.43 | 0.0092 | 0.0084 |

TABLE VI. FUTURES 3 PERFORMANCE AT DIFFERENT LAGS

| Lag | Hit Rate (%) | | RMSE | |
|---|---|---|---|---|
| | In sample | Out of sample | In sample | Out of sample |
| 1 | 70.98 | 72.45 | 0.0114 | 0.0082 |
| 2 | 71.51 | 73.31 | 0.0111 | 0.0084 |
| 3 | 71.53 | 74.05 | 0.0110 | 0.0081 |
| 4 | 73.17 | 73.43 | 0.0106 | 0.0077 |
| 5 | 73.24 | 73.58 | 0.0096 | 0.0088 |
| 6 | 73.79 | 73.80 | 0.0104 | 0.0077 |
| 7 | 75.10 | 75.03 | 0.0102 | 0.0075 |
| 8 | 75.43 | 72.69 | 0.0101 | 0.0077 |
| 9 | 74.73 | 74.29 | 0.0099 | 0.0076 |
| 10 | 75.27 | 75.28 | 0.0097 | 0.0083 |
| 11 | 75.87 | 76.51 | 0.0098 | 0.0075 |
| 12 | 76.72 | 73.80 | 0.0094 | 0.0079 |
| 13 | 76.51 | 75.15 | 0.0094 | 0.0096 |
| 14 | 76.96 | 75.40 | 0.0094 | 0.0077 |
| 15 | 77.22 | 74.05 | 0.0094 | 0.0079 |
| 16 | 77.54 | 73.68 | 0.0091 | 0.0103 |
| 17 | 77.96 | 74.42 | 0.0090 | 0.0077 |
| 18 | 77.08 | 74.17 | 0.0089 | 0.0080 |
| 19 | 77.68 | 75.77 | 0.0088 | 0.0081 |
| 20 | 77.67 | 72.82 | 0.0088 | 0.0085 |

TABLE VII. FUTURES 4 PERFORMANCE AT DIFFERENT LAGS

| Lag | Hit Rate (%) | | RMSE | |
|---|---|---|---|---|
| | In sample | Out of sample | In sample | Out of sample |
| 1 | 70.11 | 72.32 | 0.0116 | 0.0084 |
| 2 | 70.20 | 71.71 | 0.0113 | 0.0087 |
| 3 | 70.57 | 72.94 | 0.0112 | 0.0083 |
| 4 | 72.42 | 74.05 | 0.0108 | 0.0081 |
| 5 | 73.05 | 74.29 | 0.0107 | 0.0082 |
| 6 | 72.94 | 74.66 | 0.0106 | 0.0079 |
| 7 | 74.42 | 75.65 | 0.0104 | 0.0079 |
| 8 | 74.16 | 72.94 | 0.0103 | 0.0078 |
| 9 | 74.21 | 73.68 | 0.0102 | 0.0137 |
| 10 | 74.47 | 74.91 | 0.0100 | 0.0081 |
| 11 | 75.72 | 76.26 | 0.0099 | 0.0432 |
| 12 | 74.88 | 74.66 | 0.0100 | 0.0080 |
| 13 | 75.42 | 74.29 | 0.0097 | 0.0154 |
| 14 | 75.49 | 75.65 | 0.0098 | 0.0079 |
| 15 | 75.61 | 74.17 | 0.0097 | 0.0126 |
| 16 | 76.07 | 76.14 | 0.0097 | 0.0085 |
| 17 | 76.43 | 72.69 | 0.0095 | 0.0082 |
| 18 | 76.59 | 73.80 | 0.0093 | 0.0086 |
| 19 | 76.90 | 74.91 | 0.0090 | 0.0398 |
| 20 | 77.02 | 72.82 | 0.0092 | 0.0083 |

Tables 4.7 to 4.10 show that none of the futures contracts alone as input was able to outperform the benchmark. For contract 1, even with 20 lags, the forecast was less accurate than what we obtained from spot price solely. Nonetheless, it is fair to say that the performance of futures as input is not poor either. However, the real test is whether any of these contracts (or a combination of them) will improve the results if added to the benchmark.

TABLE VIII. FUTURES 1 ADDED TO THE BENCHMARK

| Metric | Hit rate | RMSE | MSE | MAE |
|---|---|---|---|---|
| In sample | 79.18 | 0.0084 | 0.0001 | 0.0063 |
| Out of sample | 80.44 | 0.0059 | 0.0000 | 0.0046 |

*C. Multi-steps Forecasts*

The final step is to test whether the model is able to forecast more than one step ahead. The input included the lagged spot price, while the target is spot price at times *t+1, t+2, t+3*.

TABLE IX. HIT RATES FOR IN-SAMPLE AND OUT-OF-SAMPLE FOR 3 DAYS FORECAST OF SPOT PRICE

| Hit rate | t+1 | t+2 | t+3 |
|---|---|---|---|
| In-sample | 78.60 | 67.45 | 54 |
| Out-of-sample | 78.72 | 66.66 | 50 |

Table 4.16 shows that the hit rate for out of sample was acceptable for up of 2 days ahead while on average it was equal to flipping a coin for out-of-sample. Moreover, table 4.17 illustrates the results of adding one lag of futures prices 1, 2, 3, and 4 months to maturity on top of the benchmark.

TABLE X. HIT RATES FOR IN-SAMPLE FOR 3 DAYS FORECAST OF SPOT PRICE ADDING 1 LAG OF FUTURES 1, 2, 3, 4

| Hit rate | In-Sample | | |
|---|---|---|---|
| | t+1 | t+2 | t+3 |
| Futures1 | 78.35 | 68.31 | 56 |
| Futures2 | 78.81 | 67.96 | 56.67 |
| Futures3 | 78.73 | 67.72 | 55.45 |
| Futures4 | 78.94 | 68.28 | 55.35 |

TABLE XI. HIT RATES FOR OUT-OF-SAMPLE FOR 3 DAYS FORECAST OF SPOT PRICE ADDING 1 LAG OF FUTURES 1, 2, 3, 4

| Hit rate | Out-of-Sample | | |
|---|---|---|---|
| | t+1 | t+2 | t+3 |
| Futures1 | 77.50 | 66 | 53 |
| Futures2 | 78.69 | 66.78 | 52.95 |
| Futures3 | 78.97 | 66.30 | 48.59 |
| Futures4 | 78.84 | 67.28 | 49.82 |

By comparing table 4.16 and 4.17 it is clear that futures contracts 1 and 2 improved the direction forecast for out-of-sample for time *t+3* while contracts 3, and 4 did not add any information to the spot price.

## V. CONCLUSION

In this paper we presented a model for forecasting crude oil prices on the short-term. In addition, we tested the relation between crude oil futures prices and spot price, and if futures are good predictors to the spot applying nonlinear ANN model. Namely, Daily spot price for WTI and futures prices for 1, 2, 3, and 4, months to maturity was considered. Data was obtained from Energy Information Administration covering the period from 1996 to 2007. Several transformation methods was tested, we find that applying 3 days simple moving average to the original data then transform it into relative change is the best methods amongst the other means tested. Moreover, attentions was paid for finding ANN model structure, as well as discovering the optimal number of lags based on spot price solely as input, and use it for benchmark purposes. Then futures price was added to the benchmark and the performance was compared. Evidence was found in support that futures prices of crude oil WTI contain new information about oil spot price direction. Futures contracts 1, 2 have preformed better than contracts 3, 4 but the overall improvement was insignificant. , Although, it could be argued that the relation between spot and futures could be different during the day. In other words, testing with intraday data could produce different results. However, intra day data for crude oil prices is not available. Finally, our future research continue to investigate other variable which could lead to improving the short-term forecast, such as heating oil prices, interest rate, and gold prices.

## VI. REFERENCES

[1] The International Energy Agency, 2004, "Analysis of the impact of high oil prices on the global economy," Report

[2] NYMEX 2006, "A guide to energy hedging."

[3] S. Moshiri, and F. Foroutan, "Forecasting nonlinear crude oil futures prices," The Energy Journal vol. 27, pp. 81-95, 2005.

[4] S. Wang, L. Yu, and K. Lai. "Crude oil price forecasting with TEI@I methodology." Journal of Systems Science and Complexity, vol. 18(2), 145-166, 2005.

[5] M. Mittermayer, and G. Knolmayer, "Text mining systems for market response to news: A survey," 2006.

[6] W. Xie, L. Yu, L, S. Xu, and S. Wang, "A new method for crude oil price forecasting based on support vector machines," Lecture notes in computer science, V.N. Alexanderov et al., Ed. Springer, Heidelberg, pp. 444-451, 2006.

[7] P. Silvapulle, and A. Mossa, "The relation between spot and future prices: Evidence from the crude oil market," The Journal of Futures Markets, vol. 19(2), pp. 175-193, 1999.

[8] C. Brooks, G. Rew, S. Alistair and S. Ritson, S, "A trading strategy based on the lead-lag relationship between the spot index and futures contracts for the FTSE 100," International Journal of Forecasting vol. 17, pp. 31-44, 2001.

[9] E. Bopp, and S. Sitzer, "Are petroleum futures prices good predictors of cash value?" The Journal of Futures Market, pp. 705-719, 1987.

[10] K. Chan, "A further analysis of the lead-lag relationship between the cash market and stock index futures market" The Review of Financial Studies vol. 7(6), pp. 123-152, 1992.

[11] A. Coppola, Forecasting oil price movements: Exploiting the information in future market. 34, http://papers.ssrn.com/paper.taf?abstract_id=967408, 2007.

[12] S. Abosedra, and H. Baghestani, "On the predictive accuracy of crude oil futures prices," Energy Policy, pp. 1389-1393, 2004.

[13] M. Labonte, "The effect of oil shocks on the economy: A review of the empirical evidence", RL31608, 2004.

[14] A. Refenes, A., Editor, "Neural networks in the capital markets" John Wiley & Sons, New York, 1995.

[15] M. Azoff, "Neural network time series forecasting of financial markets," John Wiley & Sons, Chichester, 1994.

[16] K. Hornik, M. Stinchcombe, and H. White, "Multilayer feedforwards are universal approximators," Neural Networks 2, pp. 359-366, 1989.

[17] G. Bowden, K. James, Forecasting water resources variables using artificial neural networks, 2003.

[18] D. McNeils, "Neural networks in finance gaining predictive edge in the market," Elsevier Acadmic Press, Massachusetts, 2005.

[19] M. Smith, "Neural networks for statistical modeling," Van Nostrand Reinholo, New York, 1993.

[20] Energy Information Administration, 2007, URL: http://www.eia.doe.gov/, 2007

[21] B. Vanstone, "Trading in the Australian stock market using artificial neural networks," 2005.

[22] C. Brooks, "Introductory econometrics for finance," Cambridge University press, Cambridge, 2002.

[23] R. Neuneier, and H. Zimmermann, "How to train neural networks," Neural networks: Tricks of the trade, Genevieve, B. Orr, Klaus-Robert Müller, Ed. Spring, Berlin, pp. 273-423, 1998.

[24] R. Grothmann, Multi-agent market modeling based on neural networks, 2005.

[25] S. Haykin, "Neural networks: A comprehensive foundation," Prentice Hall, New Jersey, 1999.

[26] W. Duch, and N. Jankowski, Survey of neural transfer functions. *Neural Computing Surveys* 2, pp. 163-212. http://www.fizyka.umk.pl/publications/kmk/99ncs.pdf, 1999.